\def\year{2021}\relax
\documentclass[letterpaper]{article} 
\usepackage{aaai21}  
\usepackage{times}  
\usepackage{helvet} 
\usepackage{courier}  
\usepackage[hyphens]{url}  
\usepackage{graphicx} 
\usepackage{natbib}  
\usepackage{caption} 
\usepackage{multirow}
\usepackage{fancybox}

\title{Attention Beam: An Image Captioning Approach}
\author {
    Anubhav Shrimal,
    Tanmoy Chakraborty \\
}
\affiliations {
    Department of Computer Science \& Engineering \\
    IIIT-Delhi, India\\
    \{anubhav18033, tanmoy\}@iiitd.ac.in
}

\begin{document}

\maketitle

\begin{abstract}
The aim of image captioning is to generate textual description of a given image. Though seemingly an easy task for humans, it is challenging for machines as it requires the ability to comprehend the image (computer vision) and consequently generate a human-like description for the image (natural language understanding). In recent times, encoder-decoder based architectures have achieved state-of-the-art results for image captioning. Here, we present a heuristic of beam search on top of the encoder-decoder based architecture that gives better quality captions on three benchmark datasets: Flickr8k, Flickr30k and MS COCO.
\end{abstract}

\section{Introduction}
Image captioning is an active research area due to the large number of applications where it can be used. It provides a gateway for scene understanding where the task is not just object recognition but also to capture the relations between the objects present in the image. Convolutional Neural Networks (CNNs) are known to perform well for feature extraction in images. Long Short Term Memory Networks (LSTMs) as variant of Recurrent Neural Networks (RNNs) have shown great potential in natural language modeling and text generation tasks. The idea to combine the two into an encoder-decoder architecture for image generation was first proposed by \cite{DBLP:journals/corr/VinyalsTBE14,7298932} in which the pre-trained CNN was used to extract the latent features of an image and represent it in a reduced form which are then fed to a modified RNN coupled with the word embedding inputs and history of the RNN to generate sequence of words, i.e., caption for the image. The extension of this work \cite{xu2015show} introduced a visual attention network along with the encoder-decoder framework. The intuition was that while captioning an image, rather than looking at the complete image at once, one can  look over different regions at each time step to caption it. The objective of attention network was to provide an attention map for the image pixels at each time step of caption generation which allowed the model to look into specific regions of the image while captioning. 

We further extend the architecture mentioned above by using beam search \cite{zhou2018tree} at the time of caption generation. It helps in finding the most optimal caption that can be generated by the model instead of greedily choosing the word with best score at each decoding step. Though beam search has been previously used for image captioning \cite{ma2019image}, we show that using this simple heuristic search along with better training schemes such as teacher forcing gives better scores for different evaluation metrics such as BLEU-1,2,3,4, METEOR, CIDEr and ROUGE-L.

\noindent\shadowbox{\begin{minipage}[t]{.95\columnwidth}
Our code and dataset available at \url{https://bit.ly/2kUU4g8} and a demo video is available at \url{https://youtu.be/bO4bvjYyvQE}. A graphical user interface is also created to consume the trained model (see supplementary).
\end{minipage}}

\begin{table*}[t]
\centering
\resizebox{0.90\textwidth}{!}
{
\begin{tabular}{cc|ccccccc|}
\cline{3-9}
 & \multicolumn{1}{c|}{} & \multicolumn{7}{c|}{\textbf{Evaluation metric}} \\ \hline
\multicolumn{1}{|c|}{\textbf{Dataset}} & \multicolumn{1}{c|}{\textbf{Model}} & \multicolumn{1}{c|}{\textbf{BLEU-1}} & \multicolumn{1}{c|}{\textbf{BLEU-2}} & \multicolumn{1}{c|}{\textbf{BLEU-3}} & \multicolumn{1}{c|}{\textbf{BLEU-4}} & \multicolumn{1}{c|}{\textbf{METEOR}} & \multicolumn{1}{c|}{\textbf{CIDEr}} & \multicolumn{1}{c|}{\textbf{ROUGE-L}} \\ \hline
\multicolumn{1}{|c|}{\multirow{5}{1in}{Flickr8k \\ \cite{hodosh2013framing}}} &\citeauthor{DBLP:journals/corr/VinyalsTBE14}$^{\dagger \Sigma}$ & 63 & 41 & 27 & --- & --- & --- & --- \\
\multicolumn{1}{|c|}{} &\citeauthor{xu2015show} (Soft Attention) & 67 & 44.8 & 29.9 & 19.5 & 18.93 & --- & --- \\
\multicolumn{1}{|c|}{} &\citeauthor{xu2015show} (Hard Attention) & \textbf{67} & 45.7 & 31.4 & 21.3 & 20.3 & --- & --- \\
\multicolumn{1}{|c|}{} & Ours (Beam = 1) & 60.8 & 43 & 29.4 & 19.8 & 20.9 & 50.7 & 46.4 \\ 
\multicolumn{1}{|c|}{} & Ours (Beam = 4) & 64 & \textbf{45.8} & \textbf{32.2} & \textbf{22.3} & \textbf{21} & \textbf{55.3} & \textbf{47.1} \\ \hline
\multicolumn{1}{|c|}{\multirow{5}{1in}{Flickr30k \\ \cite{young2014image}}} &\citeauthor{DBLP:journals/corr/VinyalsTBE14}$^{\dagger \Sigma}$ & 66.3 & 42.3 & 27.7 & 18.3 & --- & --- & --- \\
\multicolumn{1}{|c|}{} &\citeauthor{xu2015show} (Soft Attention) & 66.7 & 43.4 & 28.8 & 19.1 & 18.49 & --- & --- \\
\multicolumn{1}{|c|}{} &\citeauthor{xu2015show} (Hard Attention) & 66.9 & 43.9 & 29.6 & 19.9 & 18.46 & --- & --- \\
\multicolumn{1}{|c|}{} & Ours (Beam = 1)& 65.1 & 46.4 & 32.5 & 22.7 & 20.3 & 48 & 46 \\ 
\multicolumn{1}{|c|}{} & Ours (Beam = 4)& \textbf{67.4} & \textbf{49.5} & \textbf{36} & \textbf{26} & \textbf{20.1} & \textbf{52} & \textbf{47} \\ \hline
\multicolumn{1}{|c|}{\multirow{6}{1in}{MS COCO \\ \cite{lin2014microsoft}}} &\citeauthor{DBLP:journals/corr/VinyalsTBE14}$^{\dagger \Sigma}$ & 66.6 & 46.1 & 32.9 & 24.6 & --- & --- & --- \\
\multicolumn{1}{|c|}{} &\citeauthor{xu2015show} (Soft Attention) & 70.7 & 49.2 & 34.4 & 24.3 & 23.9 & --- & --- \\
\multicolumn{1}{|c|}{} &\citeauthor{xu2015show} (Hard Attention) & 71.8 & 50.4 & 35.7 & 25 & 23.04 & --- & --- \\
\multicolumn{1}{|c|}{} &\citeauthor{ma2019image} (Beam = 3) & 70.6 & 54.0 & 40.6 & 30.5 & 25.3 & 97.1 & 52.8 \\ 
\multicolumn{1}{|c|}{} & Ours (Beam = 1) & 77.1 & 61.4 & 47.1 & 35.9 & 27.9 & 114.8 & 57.3 \\ 
\multicolumn{1}{|c|}{} & Ours (Beam = 4) & \textbf{77.9} & \textbf{62.8} & \textbf{49.7} & \textbf{39.3} & \textbf{28.7} & \textbf{120.3} & \textbf{58.5} \\ \hline
\end{tabular}
}
\caption{Performance of all the competing methods for image caption generation: -- indicates unknown metric; $\dagger$ indicates a different split; $\Sigma$ indicates an ensemble. \textit{Beam = 1} is same as not using beam search.}
\label{tab:table1}
\end{table*}

\section{Proposed Approach}
We propose an encoder-decoder attention based architecture. The encoder is a ResNet-101 model pre-trained on ImageNet dataset. We remove the final classification layer of the model to use it as a feature extractor. The decoder is an LSTM model which takes the feature vector extracted by the encoder as input along with the attention map given by the visual attention model. The attention model gives a weight between $0$ and $1$ to each pixel in the image. The weighted image along with the word embedding is fed to the LSTM model at each time step which then gives a hidden state and a predicted word for current time step. It is then used by attention and LSTM network for the next decoding step (see supplementary architecture diagram). 
We use soft attention where the weights of the pixels add up to $1$. If there are $P$ pixels in our encoded image, then at each time step $t$, 
$
    \sum_{p}^{P} \alpha_{p, t} = 1
$,
where $\alpha_{p, t}$ denotes the probability or importance of pixel $p$ at time step $t$. The other attention mechanism is to use hard attention in which we choose to just sample some pixels from a distribution defined by $\alpha$. However, it is non-deterministic and non-stochastic. It gives only marginal improvements as compared to soft attention.

The following optimizations and heuristics are applied in the proposed model:

\begin{itemize}
    \item Doubly Stochastic Regularization loss function is used for the attention network. The motivation is to encourage the weights at a single pixel $p$ to sum to 1 across all time steps $T$ so that the model to attends to every pixel over the course of generating the entire sequence:
    $\sum_{t}^{T} \alpha_{p, t} \approx 1$.
    \item Fine-tune the final layers of ResNet-101 with a smaller learning rate for the purpose of image captioning as it is originally trained for image classification on ImageNet.
    \item Use Teacher Forcing to train the decoder in which the ground-truth captions are used as input to the decoder at each time step instead of the word predicted in the previous time step. This speeds up the training time by a significant margin.
    \item Beam search for better captions. A beam width $k$, (in our case $k=4$), is chosen. The algorithm selects the word sequence which has the highest cumulative score of all the words in its sequence as the caption (see supplementary).
\end{itemize}

\section{Results}
\begin{figure}[t]
    \centering
    \includegraphics[width=0.45 \textwidth]{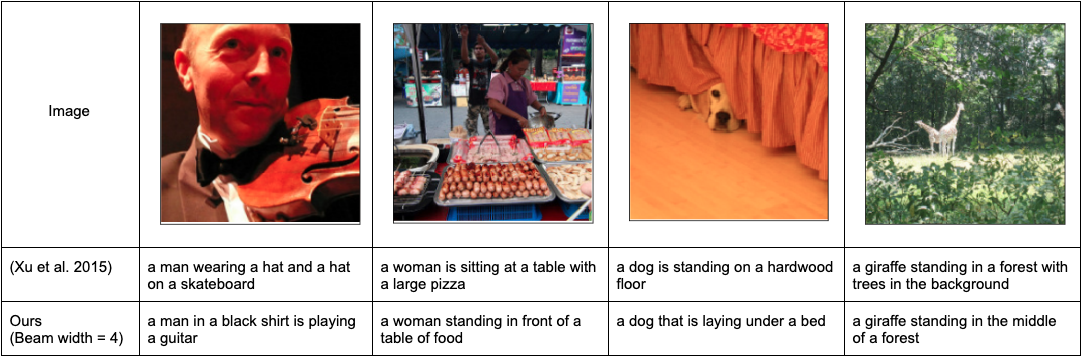}
    \caption{Captions generated by different competing methods.}
    \label{fig:compare_captions}
\end{figure}

\textbf{Data:} The experiments are performed using three benchmark datasets -- Flickr8k, Flickr30k and MS COCO, which have 8,000, 30,000 and 82,783 images, respectively. Due to the unavailability of standardized splits for Flickr30k and MS COCO, we use the splits provided in \cite{7298932}. \\
\textbf{Quantitative Analysis:} We use BLEU-1,2,3,4, METEOR, CIDEr and ROUGE-L as our evaluation metric (see supplementary for formulae). The results with various baselines are shown in Table \ref{tab:table1}. Beam search is also used by \cite{ma2019image}, but our model gives better results due to the other optimizations and heuristics in the training step.\\
\textbf{Qualitative Analysis:}  Figure \ref{fig:compare_captions} shows captions generated by different competing methods. We also compare captions generated with and without beam search, low CIDEr score captions, and visualise the attention network weights (see supplementary).

\section{Conclusion}
We proposed beam search heuristic for better caption generation for images on three benchmark datasets which shows that it beats the state-of-the-art approach. The heuristic can be applied to any given image captioning model as well as other language modeling tasks.

{\fontsize{9.0pt}{10.0pt} \selectfont \bibliography{ref.bib}}
\end{document}


\maketitle

\section{Proposed Approach}
\subsection{Architecture}
\begin{figure}[ht]
    \centering
    \includegraphics[width=0.45 \textwidth]{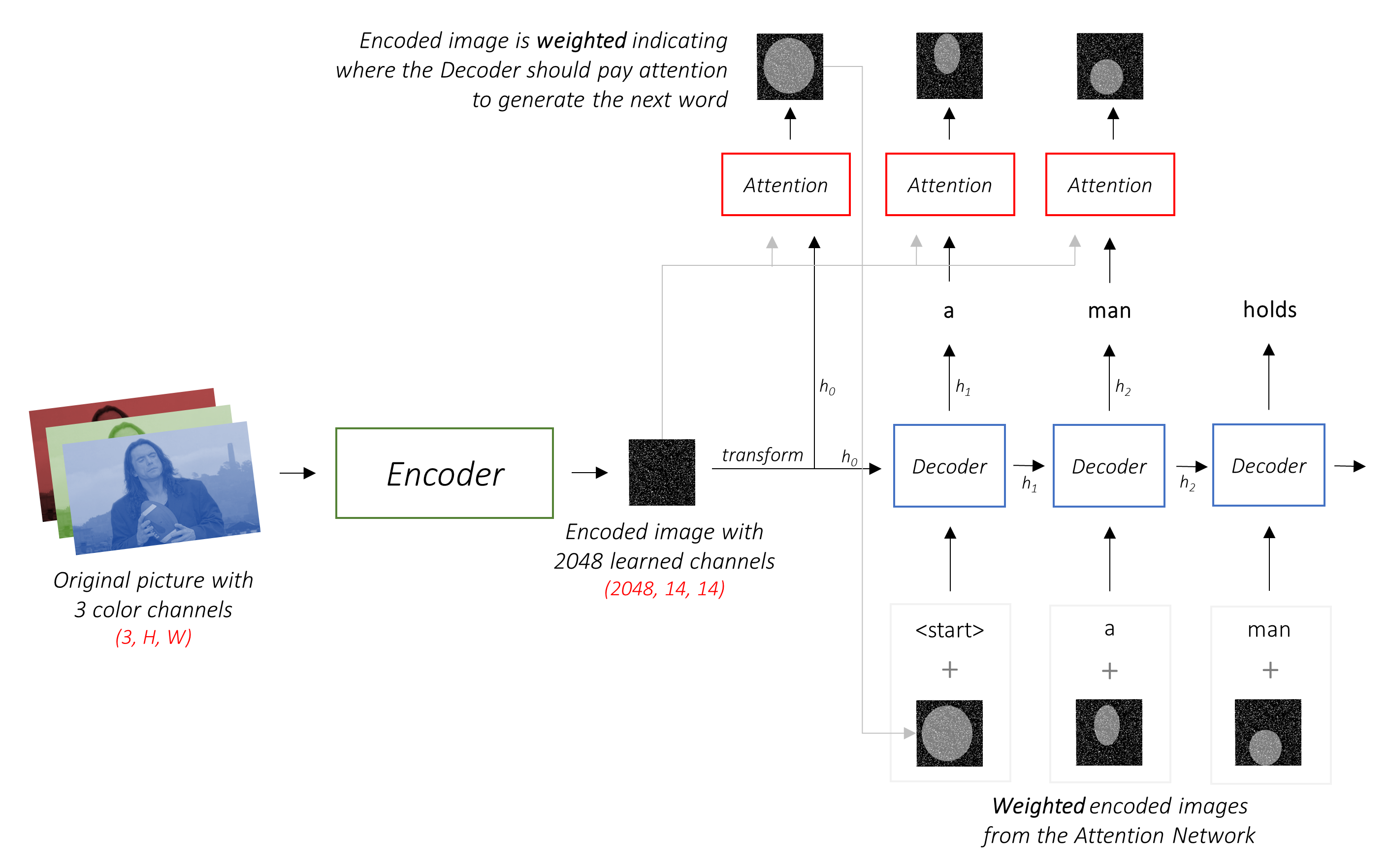}
    \caption{Model architecture.}
    \label{fig:proposed_model}
\end{figure}

Figure \ref{fig:proposed_model} shows our encoder-decoder architecture along with the attention network. The complete model is trained end to end and does not use any ensemble techniques. 

\textbf{Encoder} is a pretrained ResNet-101\cite{he2016deep} whose classification layer has been pruned to get the image feature vector. We also fine tune convolutional blocks 2 through 4 in the ResNet with a smaller learning rate of $1e-4$ to better adapt for image captioning task. 

\textbf{Attention network} takes in input the encoded image feature vector and decoder hidden state of the same dimension, these are then added and passed through a ReLU activation function followed by a linear layer which gives an output of dimension $1$ over which softmax is applied to generate the attention weights for each pixel in the image. 
We use soft attention where the weights of the pixels add up to $1$. If there are $P$ pixels in our encoded image, then at each time step $t$, 
$
    \sum_{p}^{P} \alpha_{p, t} = 1
$,
where $\alpha_{p, t}$ denotes the probability or importance of pixel $p$ at time step $t$.

\textbf{Decoder} is a Long Short Term Memory (LSTM) newtwork which takes in the flattened feature vector, initialized hidden state and the $<start>$ symbol embedding along with the attention-weighted encoding to generate the new hidden state and predict the next word in sequence along with a new hidden state.
The attention model gives a weight between $0$ and $1$ to each pixel in the image. The weighted image along with the word embedding is fed to the LSTM model at each time step which then gives a hidden state and a predicted word for current time step. It is then used by attention and LSTM network for the next decoding step. 

\begin{figure*}[ht]
    \centering
    \includegraphics[width=0.75 \textwidth]{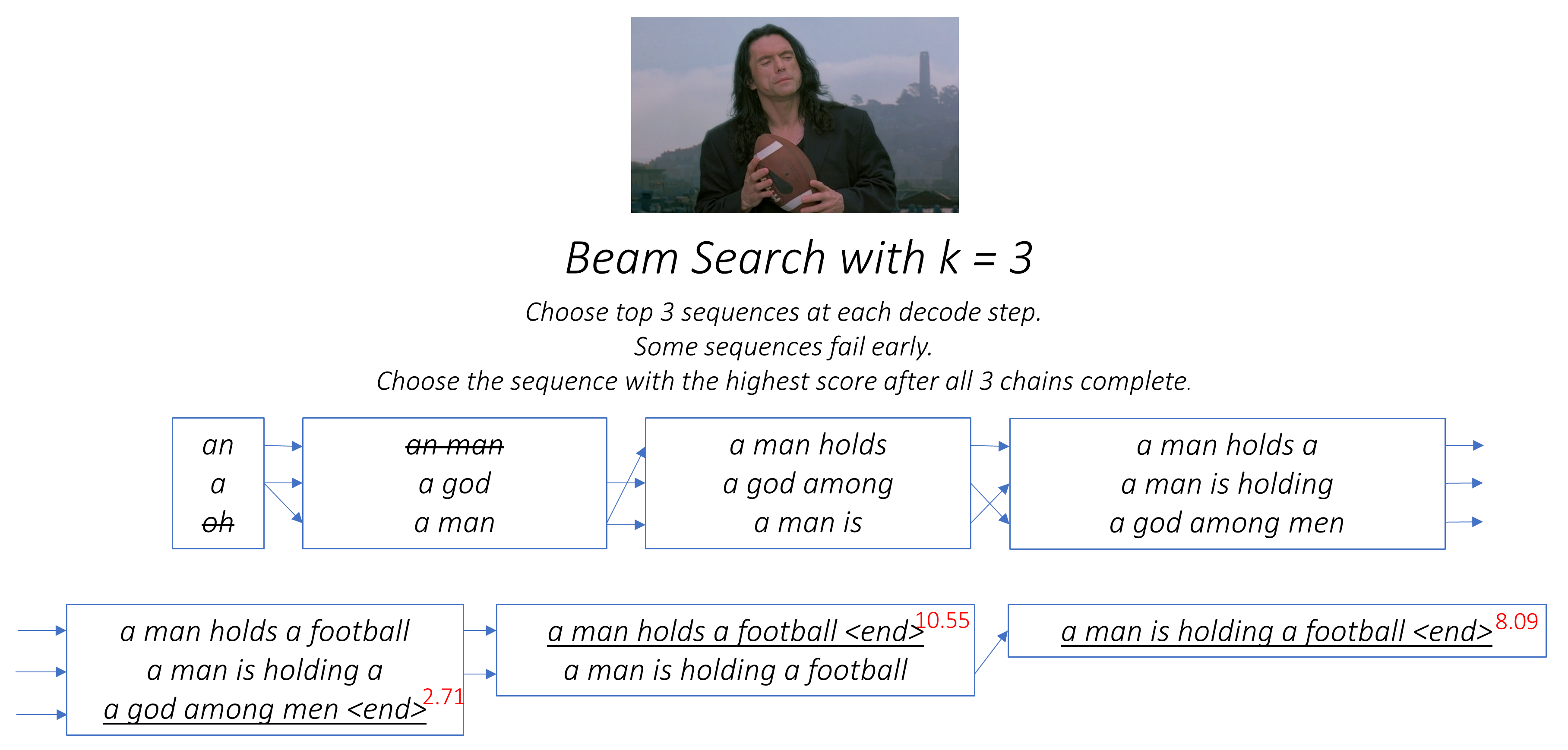}
    \caption{Beam search example with beam width $k$ = 3.}
    \label{fig:beam_search}
\end{figure*}

\subsection{Beam Search}
Figure \ref{fig:beam_search} shows an example of how beam search prunes with a beam width $k = 3$. The algorithm selects the word sequence which has the highest cumulative score of all the words in its sequence as the caption. Beam width $k = 1$ performs greedy search, that is, taking the max probability word at each decoding step. Large $k$ value gives a good chance for better caption generation, but takes more space and time. Whereas, small $k$ value is faster, but does not give good results. The following steps are followed to generate caption using beam search:

\begin{itemize}
    \item At first decoding step, top $k$ probability words are selected from the softmax prediction.
    \item At the next time step, generate $k$ words for each of the previous $k$ words.
    \item Choose the top $k$ [first word, second word] combinations considering additive scores of both the words.
    \item Repeat the process until $\langle end \rangle$ token is found and then select the caption with highest additive score of individual words (see Figure \ref{fig:beam_search}).
\end{itemize}

\begin{figure}[ht]
\centering
\begin{subfigure}{.46\textwidth}
  \centering
  \includegraphics[width=0.98\linewidth]{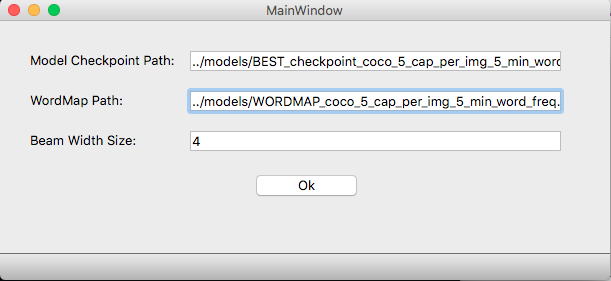}  
  \caption{first screen takes saved model, wordmap path and beam width as input}
  \label{fig:sub-first}
\end{subfigure}
\begin{subfigure}{.23\textwidth}
  \centering
  \includegraphics[width=0.95\linewidth, height=3cm]{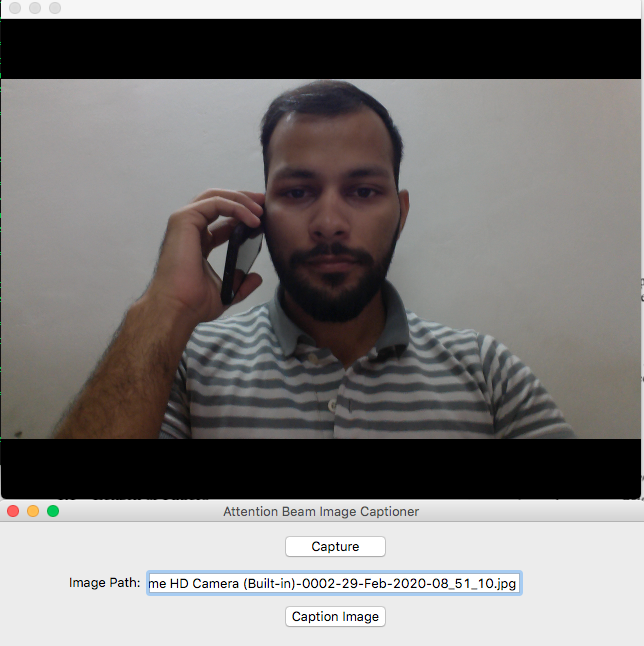}  
  \caption{second screen with camera video feed}
  \label{fig:sub-second}
\end{subfigure}
\begin{subfigure}{.23\textwidth}
  \centering
  \includegraphics[width=0.95\linewidth, height=3.3cm]{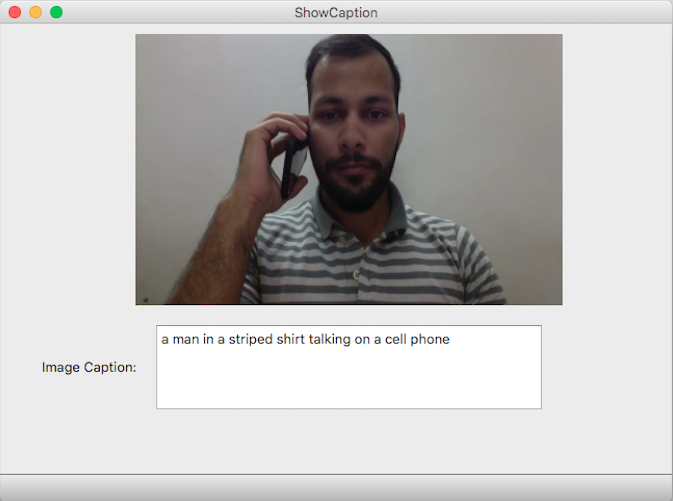}  
  \caption{third screen displays the caption generated for the captured image}
  \label{fig:sub-third}
\end{subfigure}
\caption{Graphical User interfaces for Attention Beam Image Captioning system.}
\label{fig:ui}
\end{figure}

\section{Graphical User Interface}
The proposed model can be consumed using the graphical user interface as shown in Figure \ref{fig:ui} which has the following interactions: 
\begin{itemize}
    \item Allows the user to use their trained model, word-map and give a beam width parameter ($k$) value.
    \item Allows to capture an image from the video feed, save the captured image.
    \item Generate and display the caption for image present at the specified path on the system.
\end{itemize}
A demo video on how to use the given user interface is available at \url{https://youtu.be/bO4bvjYyvQE}.

\section{Evaluation Metric Formulae}
\label{section:eval_formulae}
Following are the formulae for different evaluation metrics. Here, $a$ is the candidate sentence or generated by the model, $b$ is the set of reference sentences or ground truth, $w_n$ is n-gram, $c_x(y_n)$ is the count of n-gram $y_n$ in sentence $x$
\begin{enumerate}
    \item BLEU \cite{papineni2002bleu}(BiLingual Evaluation Understudy): It is based on n-gram precision and is geometric mean of n-gram scores from $BLEU_1$ to $BLEU_4$.
    \begin{equation}
    \begin{split}
    x &= c_{a}(w_{n}) \\
    y &= max_{j=1..|b|}c_{bj}(w_{n}) \\
    BLEU_n(a, b) &= \frac{\sum _{(w_{n}\in a)}(min (x, y))}{\sum _{(w_{n}\in a)} x}
    \end{split}
    \end{equation}
    
    \item ROUGE \cite{lin2004rouge}(Recall Oriented Understudy of Gisting Evaluation): It is based on n-gram recall.
    \begin{equation}
    \begin{split}
    x &= c_{a}(w_{n}) \\
    y &= c_{bj}(w_{n}) \\
    ROUGE_{n}(a,b) &= \frac{\sum _{j=1..|b|} \sum_{(w_{n}\in b_{j}}(min (x, y))}{\sum _{j=1..|b|} \sum_{(w_{n}\in b_{j})} y}
    \end{split}
    \end{equation}
    
    \item CIDEr \cite{vedantam2015cider}(Consensus-based Image Description Evaluation): It gives more weightage to important n-grams and is based on higher correlation with human consensus scores. Here, $g^{n}(x)$ is a vector formed by TF-IDF scores of all n-grams in $x$.
    \begin{equation}
    \begin{split}
    CIDEr_{n}(a,b) &= \frac{1}{|b|} \sum _{j=1..|b|} \frac{g^{n}(a)*g^{n}(b_{j})}{||g^{n}(a)||*||g^{n}(b_{j})||} 
    \end{split}
    \end{equation}
    
    \item METEOR \cite{banerjee2005meteor}(Metric for Evaluation of Translation with Explicit Ordering): An alignment between a and b is computed. It uses unigram precision and unigram recall. It gives smoother penalization for different ordering of chunks and is also based on higher correlation with human consensus scores. If $x$ is the number of set of unigrams adjacent in $a$ and $b_j$, $y$ is the number of matched unigrams, $P$ is unigram precision and $R$ is unigram recall, then
    \begin{equation}
    \begin{split}
    METEOR &= max_{j=1..|b|} \frac{10 PR}{R + 9P} (1- (0.5 (\frac{x}{y} )^{3}) )
    \end{split}
    \end{equation}
\end{enumerate}

\section{Results}
\begin{figure}[ht]
    \centering
    \includegraphics[width=0.45 \textwidth]{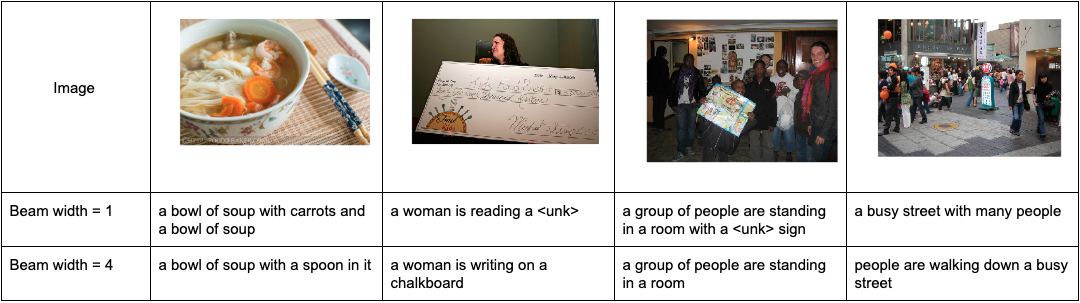}
    \caption{Low CIDEr score captions generated using beam width $k = 1$ and $k = 4$}
    \label{fig:bad_captions}
\end{figure}

\begin{figure}[ht]
\centering
    \frame{\raisebox{-\totalheight}{\includegraphics[width=.35\linewidth]{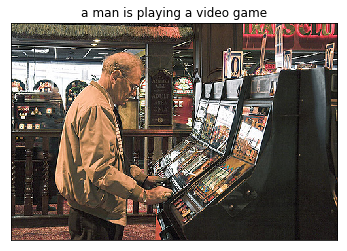}}
    \raisebox{-\totalheight}{\includegraphics[width=.64\linewidth]{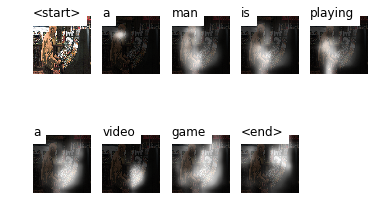}}}
    \\ 
    \frame{\raisebox{-\totalheight}{\includegraphics[width=.35\linewidth]{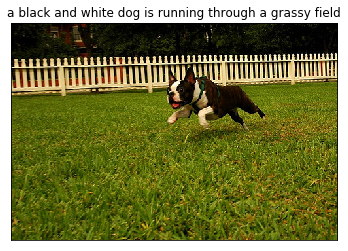}}
    \raisebox{-\totalheight}{\includegraphics[width=.64\linewidth]{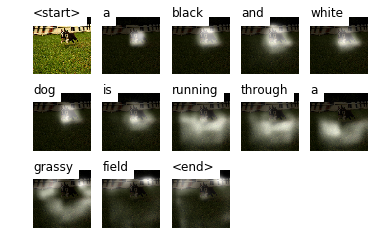}}}
    \\ 
    \frame{\raisebox{-\totalheight}{\includegraphics[width=.35\linewidth]{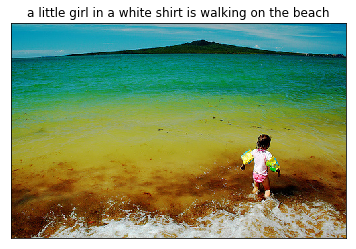}}
    \raisebox{-\totalheight}{\includegraphics[width=.64\linewidth]{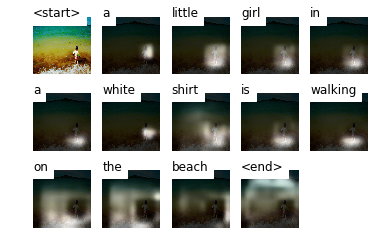}}}
\caption{Visualizing attention weights at different time step in image caption generation.}
\label{fig:attention}
\end{figure}

Figure \ref{fig:bad_captions} shows the difference in generated captions when using a beam width $k = 1$ and $k = 4$, these images had a low CIDEr score, that is, both the models produced bad captions, but it is evident from the captions that $k = 4$ produces better captions as compared to $k = 1$. Figure \ref{fig:attention} shows the weights given by the attention network to different regions of an image at different decoding time steps. This visualization gives an extra layer of interpretability to the model.

{\small \bibliography{supplementary_ref.bib}}